\def\paperID{0000}
\def\confName{ICCV}
\def\confYear{2025}

\def\paperTitle{RoboChallenge: Large-scale Real-robot Evaluation of Embodied Policies}

\def\authorBlock{
Adina Yakefu$^{2}$\thanks{Authors are listed in alphabetical order.} \; Bin Xie$^{1}$ \; Chongyang Xu$^{1}$ \; Enwen Zhang$^{1}$ \; Erjin Zhou$^{1}$ \; Fan Jia$^{1}$ \\ Haitao Yang$^{1}$ \; Haoqiang Fan$^{1}$ \; Haowei Zhang$^{1}$ \; Hongyang Peng$^{1}$ \; Jing Tan$^{1}$ \; Junwen Huang$^{1}$ \\ Kai Liu$^{1}$ \; Kaixin Liu$^{1}$ \; Kefan Gu$^{1}$ \; Qinglun Zhang$^{1}$ \; Ruitao Zhang$^{1}$ \; Saike Huang$^{1}$ \\ Shen Cheng$^{1}$ \; Shuaicheng Liu$^{1}$ \; Tiancai Wang$^{1}$ \; Tiezhen Wang$^{2}$ \; Wei Sun$^{1}$ \; Wenbin Tang$^{1}$ \\ Yajun Wei$^{1}$ \; Yang Chen$^{1}$ \; Youqiang Gui$^{1}$ \; Yucheng Zhao$^{1}$ \; Yunchao Ma$^{1}$ \; Yunfei Wei$^{1}$ \\ Yunhuan Yang$^{1}$\; Yutong Guo$^{1}$\; Ze Chen$^{1}$\; Zhengyuan Du$^{1}$\; Ziheng Zhang$^{1}$\; Ziming Liu$^{1}$\; Ziwei Yan$^{1}$ \\
$^1$ Dexmal \qquad $^2$ Hugging Face \\
Project page: \small{\url{https://robochallenge.ai}} \\
{\tt\small ~}
}

\newif\ifreview 
\newif\ifarxiv \newcommand{\arxiv}{\arxivtrue}
\newif\ifcamera 
\newif\ifrebuttal 

\arxiv 

\pdfoutput=1
\documentclass[10pt,twocolumn,letterpaper]{article}
\ifreview \usepackage[review]{cvpr} \fi
\ifarxiv \usepackage[pagenumbers]{cvpr} \fi
\ifrebuttal \usepackage[rebuttal]{cvpr} \fi
\ifcamera \usepackage{cvpr} \fi

\usepackage{graphicx}	
\usepackage{amsmath}	
\usepackage{amssymb}	
\usepackage{booktabs}
\usepackage{times}
\usepackage{microtype}
\usepackage{epsfig}
\usepackage{caption}
\usepackage{float}
\usepackage{placeins}
\usepackage{color, colortbl}
\usepackage{stfloats}
\usepackage{enumitem}
\usepackage{tabularx}
\usepackage{xstring}
\usepackage{multirow}
\usepackage{xspace}
\usepackage{url}
\usepackage{subcaption}
\usepackage{xcolor}
\usepackage[hang,flushmargin]{footmisc}

\ifcamera \usepackage[accsupp]{axessibility} \fi

% ...

\ifarxiv  \fi

\newcommand{\R}[1]{{%
    \textbf{%
        \ifstrequal{#1}{1}{\textcolor{red}{R#1}}{%
        \ifstrequal{#1}{2}{\textcolor{blue}{R#1}}{%
        \ifstrequal{#1}{3}{\textcolor{magenta}{R#1}}{%
        \ifstrequal{#1}{4}{\textcolor{teal}{R#1}}{%
                           \textcolor{cyan}{R#1}%
        }}}}%
    }%
}}

\usepackage{xr-hyper}

\makeatletter
\newcommand*{\addFileDependency}[1]{
  \typeout{(#1)}
  \@addtofilelist{#1}
  \IfFileExists{#1}{}{\typeout{No file #1.}}
}

\makeatother
\newcommand*{\myexternaldocument}[1]{
    \externaldocument{#1}
    \addFileDependency{#1.tex}
    \addFileDependency{#1.aux}
}

\definecolor{cvprblue}{rgb}{0.21,0.49,0.74}
\usepackage[pagebackref,breaklinks,colorlinks,allcolors=cvprblue]{hyperref}
\usepackage[capitalize]{cleveref}
\crefname{section}{Sec.}{Secs.}
\crefname{table}{Table}{Tables}
\crefname{figure}{Fig.}{Figs.}

\ifarxiv \crefname{appendix}{App.}{Apps.}
\else \crefname{appendix}{Suppl.}{Suppls.} \fi

\usepackage[table]{xcolor}
\usepackage{bm}

\unless\ifarxiv \myexternaldocument{_supplementary} \fi

\begin{document}
\title{\paperTitle}
\author{\authorBlock}
\maketitle

\begin{abstract}
Testing on real machines is indispensable for robotic control algorithms. In the context of learning-based algorithms, especially VLA models, demand for large-scale evaluation, i.e. testing a large number of models on a large number of tasks, is becoming increasingly urgent. However, doing this right is highly non-trivial, especially when scalability and reproducibility is taken into account.
In this report, we describe our methodology for constructing RoboChallenge, an online evaluation system to test robotic control algorithms, and our survey of recent state-of-the-art VLA models using our initial benchmark Table30.
\end{abstract}

\section{Introduction}
\label{sec:intro}

As vision-language-action models (VLA) become increasingly successful in robotic tasks~\cite{zitkovich2023rt,kim2024openvla,kim2025fine,li2024cogact,black2024pi_0,intelligence2025pi_}, the problem of benchmarking emerged.
The evaluation methodology needs to be fair enough for stable results, scalable enough to cover a wide range of tasks, and robust enough for public access. Great effort has been put into simulator-based benchmarks~\cite{mu2025robotwin,li24simpler,mees2022calvin,liu2023libero}. However, it is widely believed that a real-machine-based testing method is mandatory, since the ``real-world'' always contains factors that the digital twin cannot reproduce. This raises the problem of large-scale real-robot-based benchmarking.

\begin{figure}[t]
    \centering
    \includegraphics[width=1.0\linewidth]{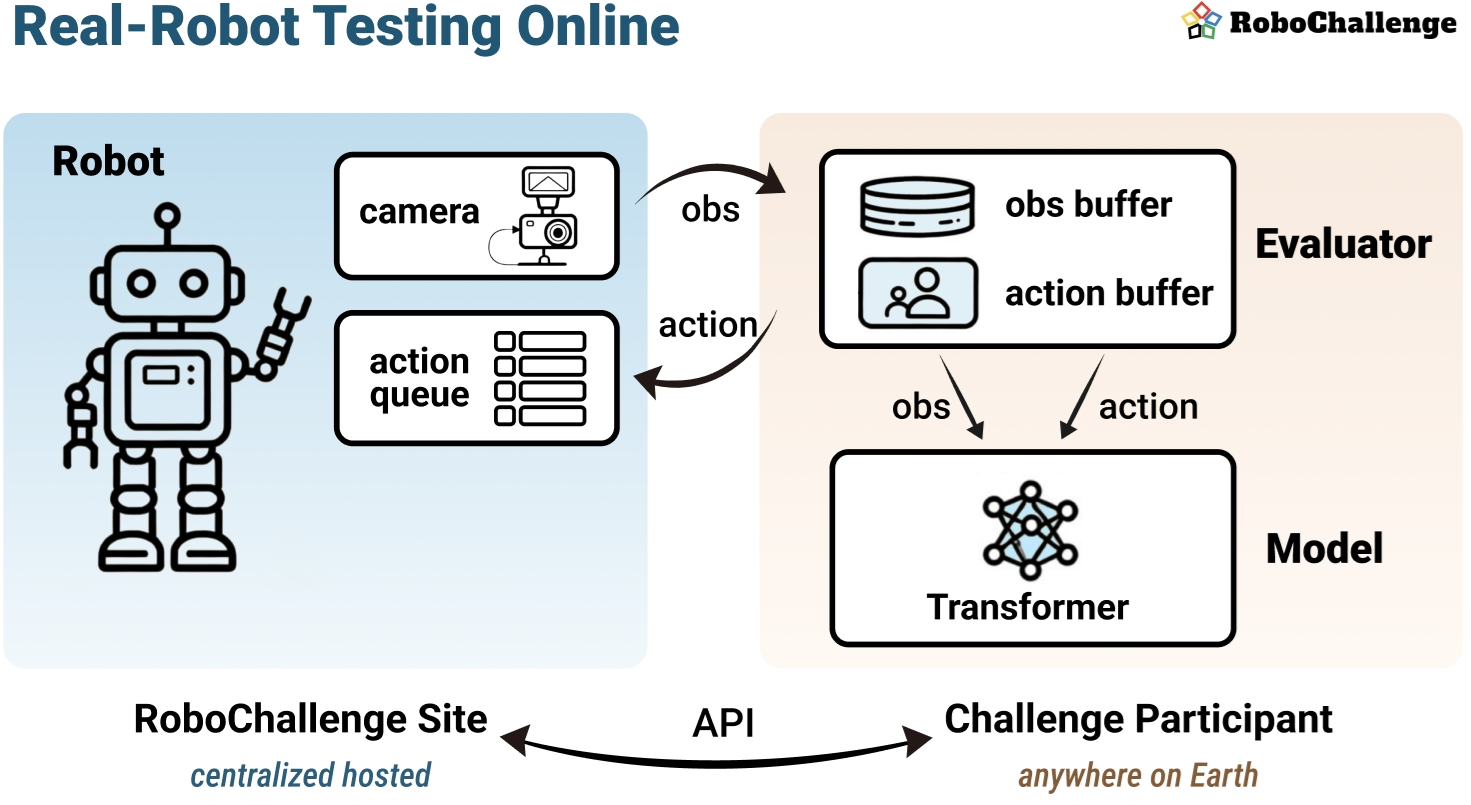}
    \caption{\label{fig:apicall}We served our robots online. A set of low-level api is formalized to provide the exact timestamp of observations and state of the action queue to enable fine-grained control. No docker images or model checkpoints are needed to be exchanged.}
\end{figure}

We approach this challenge by offering a fleet of online-hosted machines for public access. The machines are equipped with our well-engineered testing harness and we have curated a long list of tasks that can be tested on them. We term this infrastructure the RoboChallenge system. 
In contrast to existing online evaluation systems~\cite{atreya2025roboarena, zhou2025autoeval} that only host a few tasks on one or two machines, our initial offering includes a fleet of 10 machines. The machines are of four types:
\begin{itemize}
    \item \textbf{UR5}. A single 6-DOF UR5 arm with a Robotiq gripper.
    \item \textbf{Franka Panda}. A 7-DOF Franka arm, with the gripper replaced by a Robotiq one.
    \item \textbf{Cobot Magic Aloha}. Two 6-DOF arms mounted on a moving platform that mimics the Aloha system~\cite{fu2024mobile}.
    \item \textbf{ARX-5 arm}. A 6-DOF ARX-5 arm, mounted on a table.
\end{itemize}

\noindent These robots are selected because of their popularity in previous researches. They are equipped with multiple RealSense RGBD cameras as their main sensors. The user interfaces with the machines through a set of online APIs to obtain the observations and execute commands during a test.

For all tasks that can be tested on our system, we will also provide the corresponding demonstration data (up to 1000 episodes per task). Users of our system are supposed to fine-tune their model using the data and submit their evaluation requests to the system. 

\begin{figure*}[t]
    \begin{minipage}{0.48\linewidth}
    \includegraphics[width=1.0\linewidth]{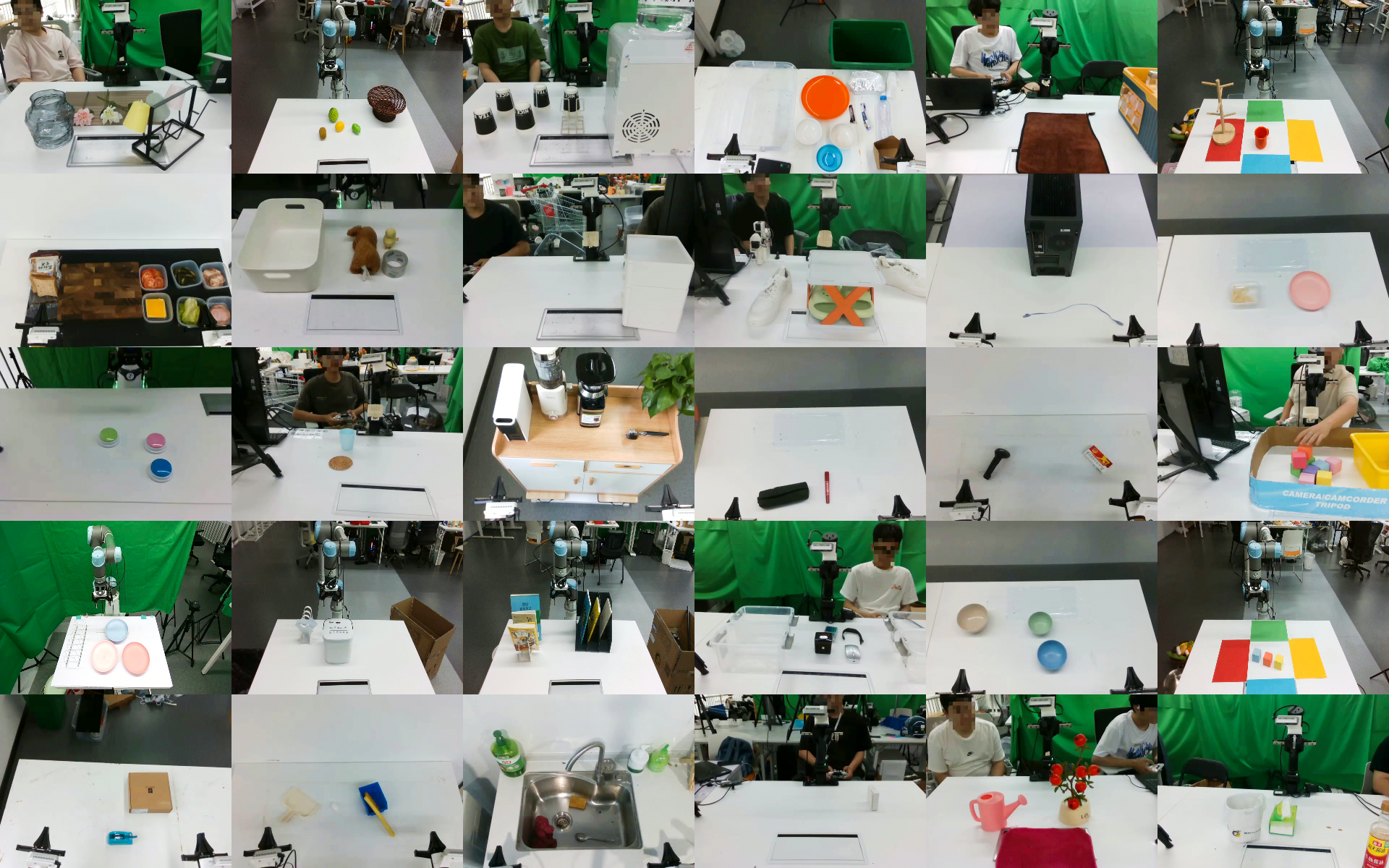}
    \end{minipage}
    \hfill
    \begin{minipage}{0.51\linewidth}
    \centering
    \includegraphics[width=1.0\linewidth]{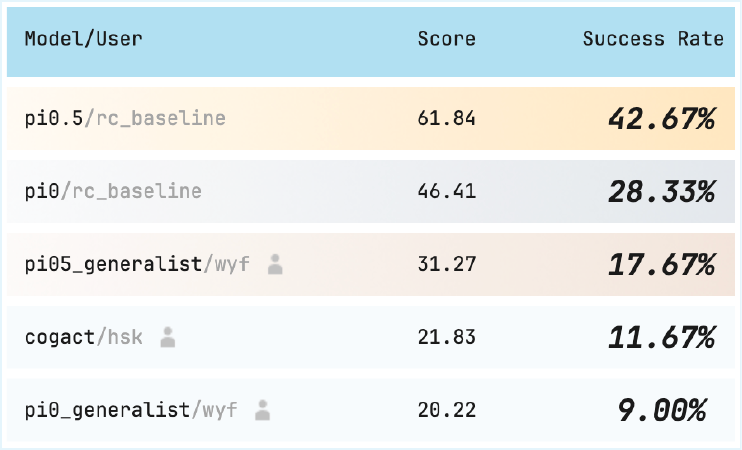}
    \end{minipage}
    \caption{\label{fig:benchmark} Left: Thumbnails of the tasks. Right: ranklist of the baseline methods. Our first benchmark is a 30-task static armed robot testing set. It challenges a variety of aspects of the learning algorithms. We measured the end-to-end task-level success rate and a score that measures the partial progress of the tasks, and see a clear distinction between the models. Models marked by \textit{rc\_baseline} are finetuned by the authors of the report. Other models are finetuned by a group of college volunteers. 
    }
    
\end{figure*}

Our initial release of the tasks includes 30 tasks tailored for testing around a fixed table. They are organized into a benchmark called Table30. This seemingly simple benchmark stresses various aspects of the learning capacity of VLA models. 
At the time of this report, five methods were tested in the 30 tasks. Two of them are implemented by our crew using the popular $\pi$ series models. The other 3 methods come from college volunteers, either using different base models or trained with a different protocol. Fig.~\ref{fig:benchmark} gives a summary of the test results. We release all the trajectories and video recordings of the robot during the test on our website.




In the following sections, we will describe the RoboChallenge system (Sec.~\ref{sec:website}), the Table30 benchmark (Sec.~\ref{sec:bmk}) and our findings (Sec.~\ref{sec:results}) in detail.

\section{RoboChallenge 
}
\label{sec:website}

Serving robots online is not as trivial as it seems. In this section, we give a detailed description and justify the design decisions of our system.

\subsection{Online Interface for Serving Robots}

The first issue is how do we expose the robots to the users that submit their algorithms. We see three major paradigms:
\begin{itemize}
    \item \textbf{Model-level submission}. The users submit the weights (and model files), and the evaluator runs the model locally. This is used in some real-robot competitions.
    \item \textbf{System-level submission}. The users submit a docker image containing the model files and inference logic, and the evaluator runs the entire system image. This is also used in some real-robot competitions.
    \item \textbf{Model API call}. The user provides an online URL that the evaluator calls to run the model. This is adopted in RoboArena. 
\end{itemize}

\noindent However, we decide to adopt \textit{none} of the methods above, for the following reasons:
\begin{itemize}
    \item \textbf{Computing}. Submitting a model and getting it running correctly on other premises is extremely tricky. The software stack (CUDA version, Python version, framework, etc.) and hardware configuration (GPU/CPU) are hard to match, and debugging is almost impossible unless full access to our machine is provided. Using docker does not solve the problem according to our experience in participating in previous competitions.
    \item \textbf{Flexibility}. We do not want to limit users to the ``stop-and-inference'' control paradigm implied by the observation-to-action mapping assumed in previous systems. Methods like Real-Time Action Chunking~\cite{black2025real} need fine-grained access to the exact timestamp of the observation and scheduling of the actions.
    \item \textbf{Accessibility}. Not everyone has a public IP, especially in the modern Internet dominated by NATs.
\end{itemize}

The method we use is called the \textbf{``remote robot''} paradigm, illustrated in Fig.~\ref{fig:apicall}.
We do not need the user to submit their model -- the model is always evaluated on the user side. We do not even run the ``glue code'' to connect the machine and the VLA model: The user is responsible for all the format conversion and post-processing of the actions. We provide low-level and fully asynchronous access of the cameras and machine to the user, enabling them to construct complex strategies for temporal alignment or ensembling.

The user access our camera by sending a \textbf{capture request}, and they will receive a set of precisely timestamped observation (RGB, depth and proprioception). At the same time, the user can post actions (with their corresponding duration time) into our \textbf{action queue}. Our robot will sequentially pop the actions in a FIFO order, and inform the user of the current length of the queue through our API. In this way, all actions sent to the queue is irrevocable, and access to the camera and the robot can be fully asynchronous.
Users never need to provide a publicly accessible API for us to call. Instead, they call ours. This makes life easier for users behind Network Address Translation (NAT).

Another often neglected set of APIs that we provide is for job scheduling. We will inform the users of the expected time that their models need to run. Before that, the users can leave their GPUs to other use, and get the model ready just minutes before the actual run. When multiple tasks are under evaluation, the user can know the exact model that they should be loading, and the progress of the whole evaluation job.

\subsection{The Robot Platforms}

There are a large number of types of robot, and we need to decide on a subset of models that are included in our system. We draw a few guidelines to make the choice:
\begin{itemize}
    \item \textbf{Durability}. The robot should operate $7\times24$ for continuous online service. It needs to be either robust enough to have a long MTBF, or cheap enough for us to replace the worn instances. The robot should not have ``undefined behavior'' within its operating space.
    \item \textbf{Popularity}. The robot should be established in the research community. Its vendor should be operating in the region of our testing site, and the production lifetime of the model should be long enough for us to purchase identical new ones for future tests.
    \item \textbf{Safety}. The robot should either have its own safety restrictions (e.g. force or torque) to avoid damage, or be weak enough so that it will not easily hurt the operator or the objects.
    \item \textbf{Performance}. The robot should support a Cyclic Position Mode of control for up to 100Hz. The repeatability should be good (at least the millimeter level) from run to run.
\end{itemize}




At the time of our initial release of the first benchmark, we provide 4 types of robot: UR5, Franka Panda, Cobot Magic Aloha and ARX-5. We will elaborate on them one by one. 
The UR5 robot is extremely durable and has a long lifespan in industrial use. We mounted a Robotiq gripper as the end effector. We use the RTDE interface for synchronous control of at most 125 Hz.
Franka Panda is also a popular choice. It has 7 DOF, so we both provide joint control mode and end-point mode. We use libfranka drivers.
The Aloha and ARX-5 systems have significantly higher failure rates, but they are much cheaper. During hardware failure, we revoke the running evaluation and resume the job after maintenance. We use the CAN drivers provided by the vendors.

In addition to robots, we also need to install sensors. Our default choice is Intel RealSense depth sensors for their wide application in robotic research. They provide both time-stamped RGB and depth streams that current VLAs may depend on.
There is always a ``main'' camera that looks down on the operating area and a ``wrist'' camera installed at the end of the arm. There will be a ``side'' camera for single-arm setups.
The robots are connected to their own workstation computers. The cameras are connected to the computer through USB cables. Dedicated software is written to collect demonstration data and conduct tests.
We will also consider torque or force sensors in the future, but at the current stage, we omit them for simplicity.

\subsection{Evaluation Protocol}

One of the major obstacles in real machine testing is the dramatic variation of the test results from run to run. In our experience, even with the same set of props, task and model, the measured success rate can change even from 0\% to 100\% or vice versa. Hence, we need a principled methodology to control the factors in the tests.

\subsubsection{Variation of Testers}
\begin{figure}
    \includegraphics[width=1.0\linewidth]{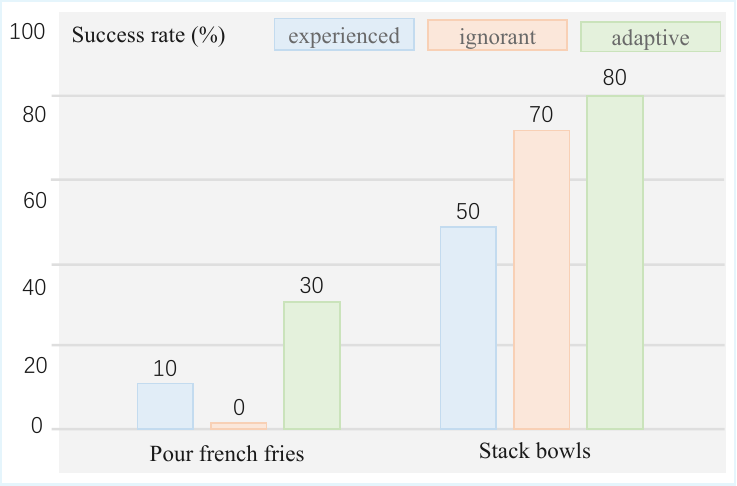}
    \caption{\label{fig:variation} Variation of success rates caused by testers. We picked two tasks and tried three different testers, each with (1) \textbf{experienced tester}: the same one that collected the training data (2) \textbf{ignorant tester}: totally fresh tester seeing the prompt and props for the first time  (3) \textbf{adaptive tester}: a tester with algorithm experience and managing to ``improve'' the success rate as much as he/she could.
    }
\end{figure}

Our first observation is the variation caused by the testers. In a real-robot test, the human tester is responsible for preparing the props (from an available pool), resetting the objects to their (usually randomized) initial status and overlooking the run. Depending on the way objects are picked and prepared, the result may vary.

To elaborate on this issue, we pick two tasks with the corresponding models and let three groups of human testers do the test:
\begin{itemize}
    \item The \textbf{experienced testers}. They are the same group of people collecting the demonstration data. They know and are told to mimic the distribution of the demonstration episodes as much as possible.
    \item The \textbf{ignorant testers}. They are told to do the test immediately after reading the task instructions. Their understanding of the task comes only from the description of the text on how the task should be prepared and their own ``common sense''.
    \item The \textbf{adaptive testers}. They are the authors of the models. They have high incentive to create a ``good'' result in the test. We observe that their placement of the objects is strategic: depending on the result of previous runs, they manipulate the position of the objects in a seemingly random way for better results in the following runs.
\end{itemize}

The results are shown in Fig.~\ref{fig:variation}. Even with a sufficient number of runs, the recorded success rate varies considerably. The adaptive testers get better results. The ignorant testers' results are more unstable: We empirically observe a strong bias on their setup in the test runs. Experienced testers made a better attempt to provide fair results, although repeating a precise ``distribution'' is inherently difficult for humans.

\begin{figure}
    \centering
    \includegraphics[width=1.0\linewidth]{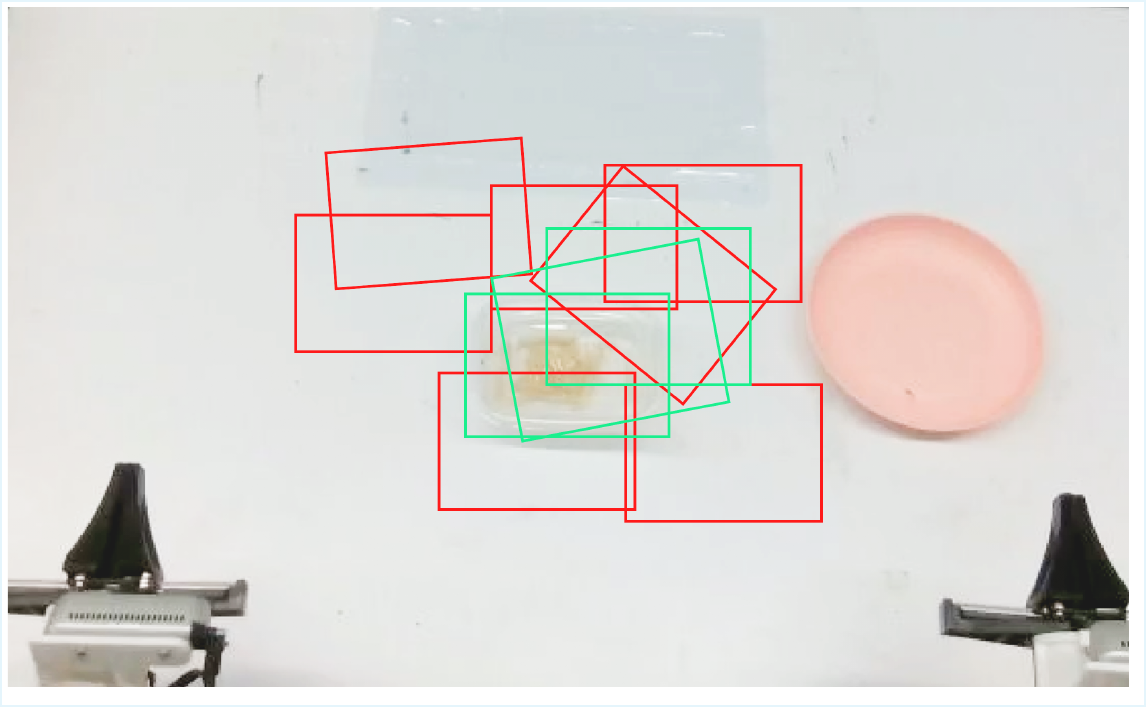}
    \caption{\label{fig:sweetspot}The ``Sweet-spot Effect''. We plot the positions of the box chosen by an ``adaptive tester'', and use green and red color to indicate a successful or failed task. The tester managed to find the location and orientation of the box that the task is more likely to succeed, and exploited this for maximal performance. This biased the test.
    }
\end{figure}

To gain more insight into how adaptive testers distort the evaluation, we observe a ``sweet spot effect''. This is exemplified in Fig.~\ref{fig:sweetspot}. As shown in the figure, there is a particular favorable set of object positions in which the task is more likely to succeed. The adaptive tester exploited these areas, resulting in seemingly improved performance. 
As for the ignorant testers, they may accidentally run into the ``sweet spots'', or ``counter-sweet spots'' that the model does not generalize. And because they are also not clear of the range the position of the objects should vary, the result becomes unstable.
The considerations above lead us to design a better protocol to do the evaluation, and in particular, a more stable method to reset the objects.

\subsubsection{Visual Task Reproduction}

\begin{figure}[t]
    \centering
    \includegraphics[width=0.99\linewidth]{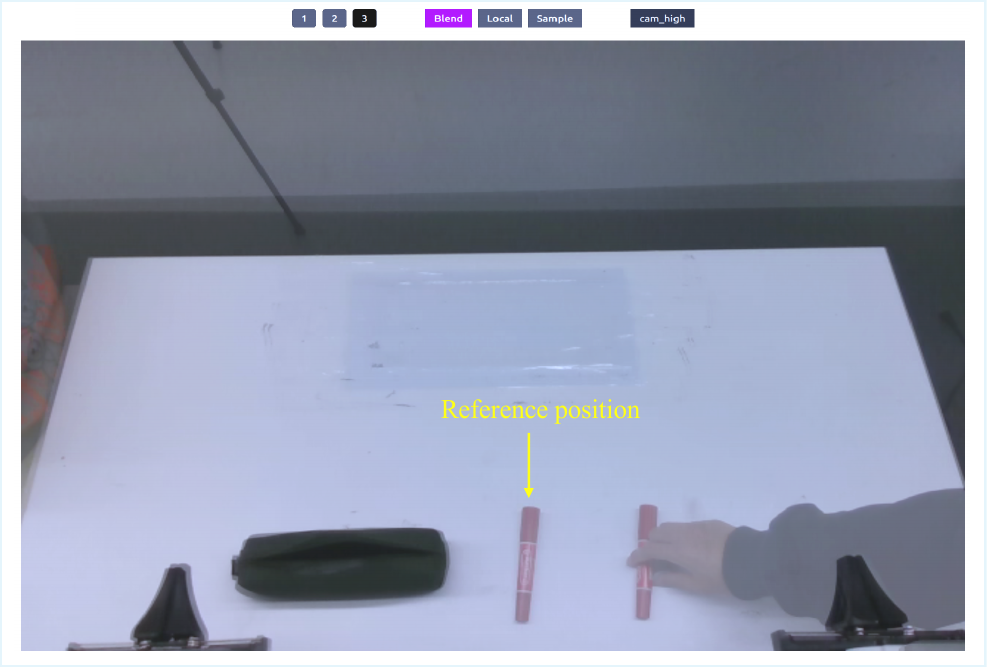}
    \caption{\label{fig:overlay}
    The tester's user interface for Visual Task Reproduction. A reference image is superimposed on the live camera stream. The tester is instructed to adjust the position of the objects and other factors so that the images match.
    }
\end{figure}

In our benchmark, we decide to control task preparation by matching \textbf{visual inputs}. We first sample a number of episodes in the demonstration data as ``reference episodes'', leaving them out for training. During each rollout, we retrieve the initial frame from one of the reference episodes and superimpose the image onto the preview stream seen by the tester (see Fig.~\ref{fig:overlay}). The tester is instructed to adjust the position of the objects until the actual input matches the reference. Also, the tester checks the consistency of other factors (e.g. position of the table, etc.).

We call this method the \textbf{controlled tester}. In this way, the initial state of the scene and objects is largely fixed across the evaluation of different models. In addition, the tester does not need a deep familiarity with the data demonstration process, making the tests scalable. Empirically, we observe that the stability of the tests using this method is even better than the ``experienced tester'' one.

\subsubsection{Background and Environmental Issues}
Although we managed to control many factors in the task setup, there are always aspects that we cannot control. The lightning condition may change from day to day. The extrinsic of the cameras is subject to drifting over a long time. We believe that precise optical-grade reproduction of the test cases is not what Embodied Intelligence should be about. We should leave the remaining factors that we do not control as an intended ``perturbation'' to the data that the model should generalize around.

\begin{figure}
    \centering
    \begin{minipage}[t]{\linewidth}
    \centering
        \begin{subfigure}[b]{\textwidth}
            \centering
            \includegraphics[width=1.0\textwidth]{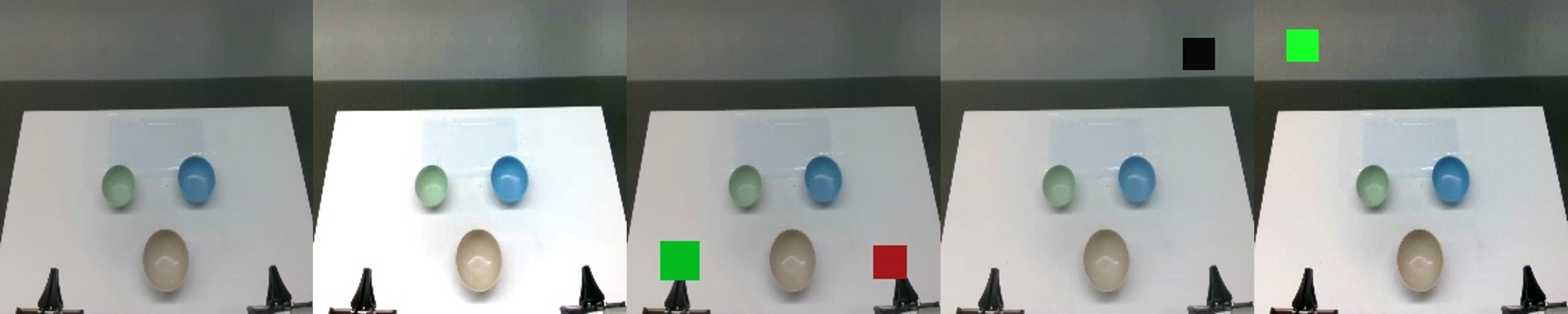}
            \caption{Input augmentation. First image on the upper left is original image.}
            \label{fig:robustness_a}
        \end{subfigure}   
        \begin{subfigure}[b]{\textwidth}
            \centering
            \includegraphics[width=1.0\textwidth]{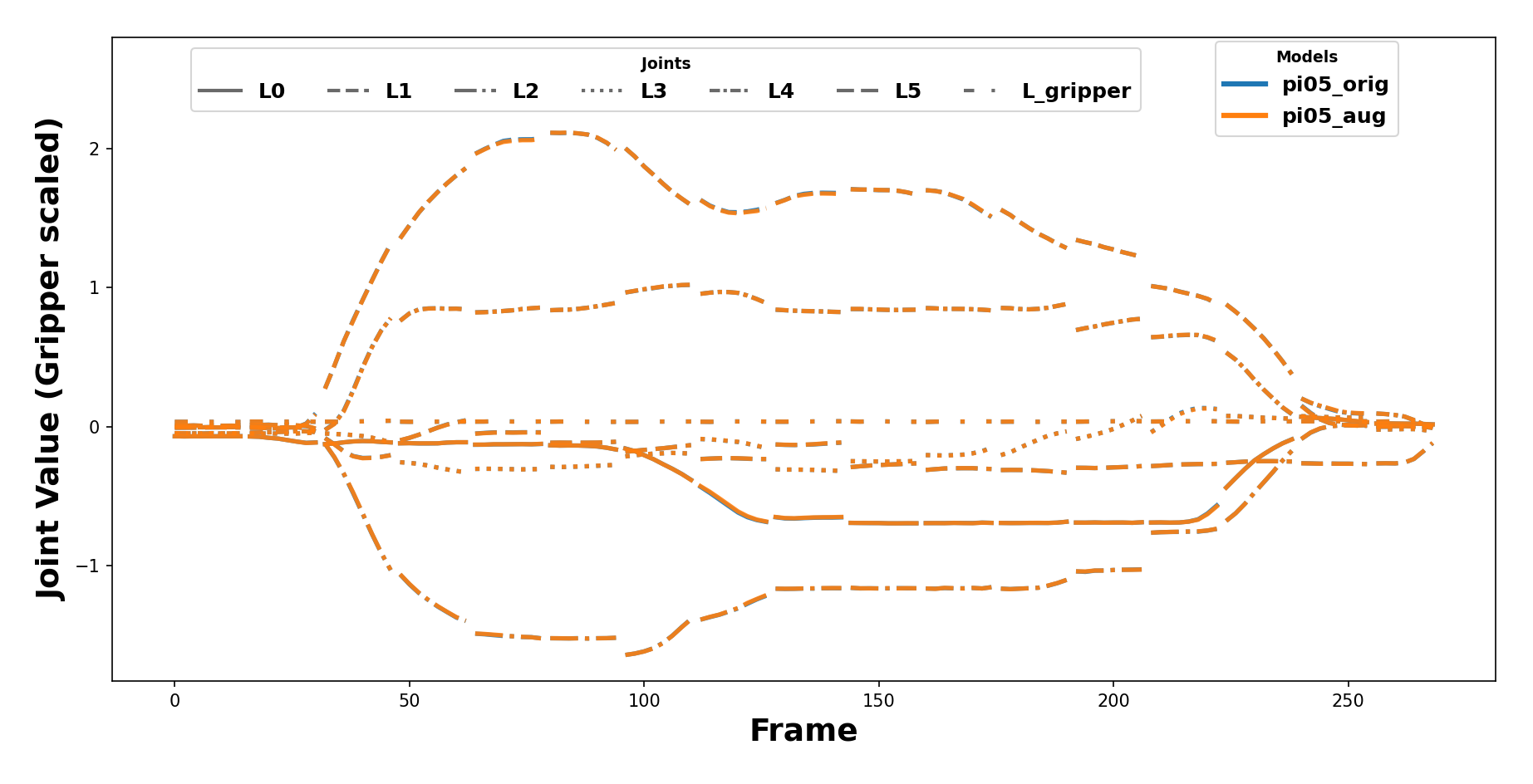}
            \caption{The outputs of VLAs with and without input augmentation.}
            \label{fig:robustness_b}
        \end{subfigure}
    \end{minipage}
    \caption{\label{fig:robustness}Robustness of VLAs. We empirical observe that background change or occlusions do not hamper nowadays VLAs. To validate this, we run the models on manually augmented or corrupted images. Consistent with our experience, the output largely matches despite change of input. This shows that the VLAs are not sensitive to these factors.
    }
\end{figure}

This claim is consistent with our experience with the models. VLAs, with billions of parameters, are inherently more robust against perturbations and distractions. To illustrate this point, we performed a proof-of-concept experiment, as shown in Fig.~\ref{fig:robustness}. We pick an input from a typical run of the model, manually corrupting or perturbing the images. The output of the model, as drawn in the Fig.~\ref{fig:robustness}~(b), remains steady. This confirms our empirical observation that the change in background or environment does not alter the test results much.

\subsubsection{Stability v.s. Fairness}

As the last missing piece of the our test protocol, we want to distinguish between two concepts:
\begin{itemize}
    \item \textbf{Stability} means the variation of the test results when \textit{one model} is evaluated on the same task for multiple times. Good stability ensures the claimed test results on one model should be reproducible by others.
    \item \textbf{Fairness} means how stable the \textit{relative order} of the models are when evaluated on one task or the same set of tasks. Good fairness ensures meaningful comparison between models.
\end{itemize}

Ideal stability implies fairness. However, in real-world this is not achievable. The rollouts conducted by the tester and the machine, however careful we control them, contains correlation (e.g. time-of-day, whether, etc). On the other hand, fairness does not imply stability, as both models can have higher or lower grade at the same time.

The approach we take in the above subsections is called the \textbf{benchmark protocol}. Its main focus is only on measuring the overall progress of individual models, i.e. stability of the results. However, we do propose an alternative protocol that focuses on fairness.
In \textbf{comparative protocol}, we implement a post-selection procedure to compare a set of models at the same time:
\begin{itemize}
    \item The tester prepares the initial state;
    \item One of the models is randomly selected and called;
    \item The tester overlooks the test without knowing which model is running.
\end{itemize}
Mathematically, we can see that the protocol is fair. The tester cannot alter the relative order of the models if one is deemed to be better than another.
We may hold competitions using this format in the future. In the current state, we only serve the ``benchmark protocol''.

\subsection{Limitation and Known Issues}
As for the machine interface, the major drawback of our inference-on-user-side approach is that we have no means to check whether the model actually run by the user matches the user's claim. The user may use a solution totally different from their submitted ``model name'', or even worse, use individually tuned models when a multi-task generalist model is expected. In theory, the user can even do human-in-the-loop cheating runs. We try to believe in the integrity of the users and encourage all users to release their models and source code so that others can reproduce the results.

One concern about the object resetting method is that as our test distribution is fixed, there is a chance that the model submissions ``overfit'' to the particular reference test cases. In practice, we have not observed this overfitting.

\section{Table30 Benchmark
}
\label{sec:bmk}

After describing our testing system, we are now at the stage to introduce our first benchmark, called Table30.

\subsection{The Tasks}
\label{subsec:tasks}

The full list of tasks is in Tab.~\ref{tab:tasks}. All tasks are executed on the table, or around a table. This gives the name of the benchmark. 
Simple as it first seems, these tasks measure a diverse set of properties that a general robot control algorithm should have. We elaborate on a few of them:

\begin{itemize}
    \item \textbf{Precise 3D Localization}: The robot need to grab or place the object at an accurate 3D position. This stress the fine-grained spatial understanding ability of the model.
    \item \textbf{Occlusion and Multi-view}: At some point, the object or the robot may be occluded in the main view. The model needs to utilize information from multiple cameras.
    \item \textbf{Temporal Dependence}: The same set of observations may appear at different stages in the task (e.g. goto a place and go back). The model need to memorize its progress.
    \item \textbf{Multi-stage and Long Horizon Tasks}: Many tasks involve doing the job in sequential steps, or repeating a skill multiple times. For a complete success of the task, the model needs to have high success rate on individual actions.
    \item \textbf{Recognizing the Object}: The robot is asked to distinguish between the objects it sees. The model should generate different actions depending on the identity of the object.
    \item \textbf{Using both Arms}: Some of the tasks require using both arms to manipulate the object, or deciding which arm to use depending on the object.
    \item \textbf{Soft Bodies}: The robot needs to deal with soft materials like towel or papers. The algorithm need to generalize to non-rigid and deforming objects.
\end{itemize}

All these factors contribute to the seemingly weird fact that even the most SOTA base model fails to achieve an overall high success rate, as shown in Fig.~\ref{fig:benchmark}. So we argue that our benchmark is a ``necessity test'' for a method in the pursuit of general robotics. 
As our evaluations accumulate, we may find more trends in the testing results. We believe the problems above are all valuable on their own rights, and we provide subset rankings in our system for further inspection.

\begin{table*}[h]
    \centering
    \begin{tabular}{lll}
        Task Name & Machine Type & Description\\
        \hline
arrange flowers & ARX5 & Pick up three flowers on the table and insert them into the vase\\
arrange fruits in basket & UR5e & Put four fruits into the basket on the table  \\
arrange paper cups & ARX5 & Stack five paper cups and put them into a shelf \\
clean dining table & ALOHA & Place trash and the dishes on the table into trash bin and basket \\
fold dishcloth & ARX5 & Fold the dishcloth two times and put it on the side \\
hang toothbrush cup & UR5 & Hang a cup on the cup holder \\
make vegetarian sandwich & ALOHA & Make a vegetable sandwich \\
open the drawer & ARX5 & Open the drawer \\
place shoes on rack & ARX5 & Place a pair of shoes on the shoe rack \\
plug in network cable & ALOHA & Insert two RJ45 connectors  into the socket \\
pour fries into plate & ALOHA & Open the box and pour the fries onto the plate \\
press three buttons & Franka & Press the pink, blue, and green buttons in sequence \\
put cup on coaster & ARX5 & Place the cup on the coaster \\
put opener in drawer & ALOHA & Place the can opener into the right-hand drawer \\
put pen into pencilcase & ALOHA & Place the pen on the table into the pencil case \\
scan  QR code & ALOHA & Scan the QR code on the medicine box using the scanner \\
search green boxes & ARX5 & Pick all green boxes in the pile into the yellow box \\
set the plates & UR5 & Place the three plates onto the plate rack one by one \\
shred scrap paper & UR5 & Stuff the paper into the shredder\\
sort books & UR5 & Place three books into corresponding position on shelf \\
sort electronic products & ARX5 & Put the four electronic products into four baskets \\
stack bowls & ALOHA & Stack three bowls together \\
stack color blocks & UR5 & Stack the yellow block on top of the orange block \\
stick transparent tape to box & ALOHA & Tear off a piece of clear tape and stick it onto the box \\
sweep the rubbish & ALOHA & Sweep the trash into the dustpan using a broom \\
move objects into box & Franka & Place all the clutter on the desk into the white basket \\
turn on faucet & ALOHA & Grasp the faucet switch and turn it on \\
turn on light switch & ARX5 & Turn on the light switch \\
water potted plant & ARX5 & Water the potted plant using the kettle \\
wipe the table & ARX5 & Grab a tissue to wipe the stains on a table and discard it \\
        
    \end{tabular}
    \caption{\label{tab:tasks} The task list, ordered alphabetically.
    }
\end{table*}

\subsection{Grading Protocol}

In our evaluations, we find that a single ``success rate'' metric is not sufficient for a fine-grained analysis. For hard tasks, a model may make good progress, but still fail in the very last step. For easy tasks, we also want the robot to complete it with the minimal number of retries or imperfections. So we defined a \textbf{progress score} to better describe robot behavior.

For a task, we divide the task into multiple stages. Each stage is assigned a certain number of progress points. After completion of each stage, the corresponding points are rewarded. A stage can be marked as ``not critical'', which means that we will mark the task as successfully executed even if this stage is not completed. An example for ``opening the drawer'' is shown in the following table:

\begin{center}
\begin{tabular}{l|c|c}
Stage & Points & Critical \\
  \hline
 Arm reaches the drawer region   &  2  & yes\\
 Grabber is rotated towards the handle &  3  & yes\\
 The drawer is pulled open & 4  & yes\\
 Arm goes back to its original position & 1  & no\\
\end{tabular}
\end{center}

During the execution of the task, a robot may ``retry'' a stage, for example, attempting to pick up an object but grabbing on the wrong position and quickly going back to the object for a second grasp. We will deduct the progress score by $0.5$ for each retry. If the progress score of a stage goes to a negative value or if the number of failed successive retries exceeds 4, we will terminate the rollout to save testing time. 
For each evaluation, the total number of progress points is 10. We make 10 rollouts for each task. So, the total progress score of a task is 100.
There may well be a task that succeeds but has a very low progress score, if the number of retries is huge. On the other hand, a task can fail at a high score if the failure occurs at the last step. So, success rate and progress score measure different aspects of the runs.

\subsection{Designing Tasks that Differentiates}

We want to stress that the selection of tasks is not arbitrary. During the design of the tasks, we keep in mind the following principles:

\begin{itemize}
    \item \textbf{Coverage of Level of Difficulties}. The difficulties of the task should range from ``very easy'' to ``difficult''. The ability of current models vary by a broad range, and we want all models to find their rooms of improvement in our benchmark.
    \item \textbf{Coverage of Algorithmic Challenges}. The tasks should cover many different aspects of the difficulties that a VLA will encounter in robotics. We give a list of the problems in Sec.~\ref{subsec:tasks}, and we want the tasks to have a fair distribution among these aspects.
    \item \textbf{Coverage of Real Life}. We want the tasks to span across a diverse range of scenarios in people's everyday life. The tasks are sampled from people's actions in their house, in a restaurant, in a workplace, or even some toy tasks that one may learn to solve during childhood.
    \item \textbf{Keep it Simple}. Given all of the requirements above, we want to keep the tasks in their simplest forms. They should all look ``trivial'' enough that a human can do these without any prior knowledge or training. We want this benchmark to be the necessary condition for general robotics.
\end{itemize}

\begin{figure}[t]
    \centering
    \includegraphics[width=1.0\linewidth]{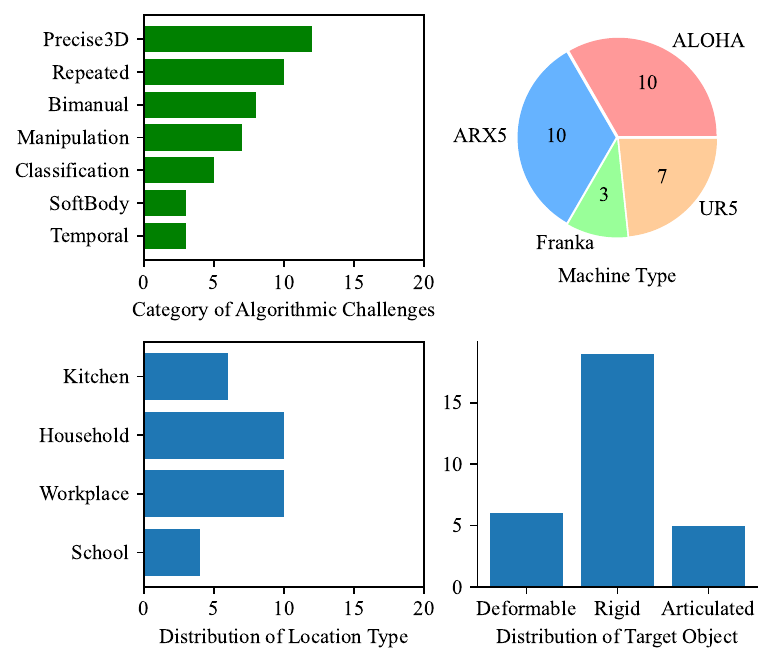}
    \caption{Distribution of our tasks. We tag our tasks either (1) by the difficulties encountered by a VLA solution (2) by the type of robot (3) by the intended location of the task scenario (4) by the property of the main target object. It shows good diversity and coverage.
        \label{fig:task_dist}
    }
\end{figure}

The outcome of following the principles above is that if an algorithm makes fundamental progress on previous solutions, we should see a clear gap in overall performance. Either it ``unlocks'' new learning abilities, enables new working scenarios, or expands its generalization ability to new object types, it will be awarded incremental points for its progress.

We plot the distribution of our tasks under various classification methods in Fig.~\ref{fig:task_dist} to show the diversity of the tasks.
There may well be another set of 30 tasks that meets all the criteria above. However, as the first work of its kind, we believe our selection is typical enough for our benchmarking purpose.

\section{Results on Table30}
\label{sec:results}

\subsection{Methods and Results}

\begin{figure}[t]
\centering
\includegraphics[width=0.85\linewidth]{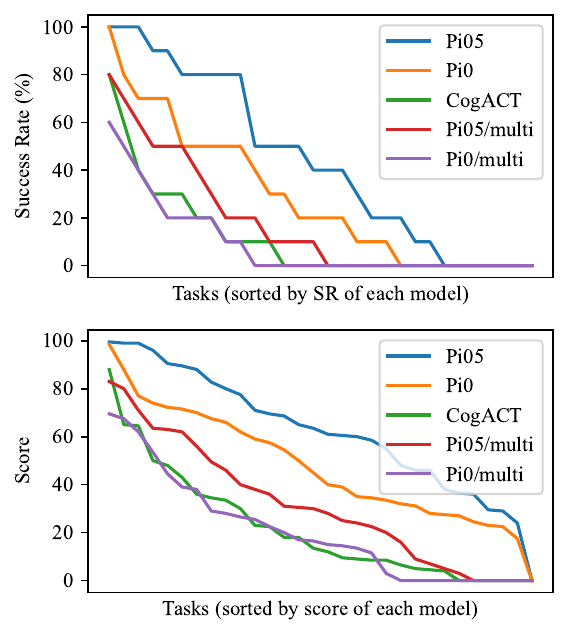}
\caption{Distribution of SR and scores. We sort the tasks by SR or score for each model to obtain the cumulative distribution.\label{fig:all_scores_plot}}
\end{figure}

\begin{figure*}
\centering
\begin{tabular}{l|cc|cc|cc|cc|cc}
\multirow{2}{*}{\textbf{Task}} & \multicolumn{2}{c|}{Pi05} & \multicolumn{2}{c|}{Pi0} & \multicolumn{2}{c|}{CogACT} & \multicolumn{2}{c|}{Pi05/multi} & \multicolumn{2}{c}{Pi0/multi}\\
\cline{2-11}
 & SR & score & SR & score & SR & score & SR & score & SR & score \\
\hline
\input{tables/scores}
\end{tabular}
\caption{\label{fig:taskresults}Results of the models on our benchmark. The color is used to indicate the best result in the row and how far it is from a perfect run. The methods with /multi suffix follows the Generalist protocol. Others follow the Task-specific protocol.}
\end{figure*}

As our initial survey, we tested four popular open source VLA algorithms:
\begin{itemize}
\item \bm{$\pi_0$}, a method open sourced by Physical Intelligence;
\item \bm{$\pi_{0.5}$}, the successor of $\pi_0$;
\item \textbf{CogACT}, an open source VLA model from Microsoft;
\item \textbf{OpenVLA/OFT}, a method derived from OpenVLA.
\end{itemize}

The models are tested in two settings. The first setting is called the \textbf{Task-specific} setting. 
The model is separately trained using all the demonstration data provided in the for each task. As there is a good number of episodes, the training typically takes 1 day on an 8-GPU machine. The second setting is called the \textbf{Generalist} setting. We sample a few samples (about 50) from each task and mix them to train a model. In our implementation, we only mix data from the same type of machine, so the model is actually a ``machine generalist''. 
The result is shown in Fig.~\ref{fig:taskresults}. We show the averaged success rate and progress score of all models. We also list the performance of individual tasks. All results can also be viewed in \url{https://robochallenge.ai/}.

\subsection{Analysis of the Models}
The first clear trend is that strong models perform significantly better. As shown in Fig.~\ref{fig:all_scores_plot}, there is a clear difference between the models. The $\pi_{0.5}$ model (finetuned) dominates all other models at all percentiles of success rate or progress score. What is more, when we look at cumulative distribution of SR of the models, we see that they all have a similar slope, meaning that the distribution of the difficulties of the tasks is rather even. We would expect future stronger models to go further in the ``upper right'' direction.

It is impressive that $\pi_{0.5}$ has a fair performance even when only a few (about 50) episodes are provided and the tasks are trained together (the \textit{Pi05/multi} entry in the figures and tables). On some tasks, this model even achieves higher scores than the task-specific finetuned model. We see this as a good indicator that the real ``generalist'' models will arrive someday.

\subsection{Analysis of the Tasks}
\begin{table}
\centering
\begin{tabular}{lccc}
\textbf{Tag} & \textbf{Tasks} & \textbf{SR} & \textbf{Score}\\
\hline
temporal &  3  &   5 &   14\\
softbody  & 3  &   8 &   27\\
precise3d & 12 &   18 &   38 \\
bimanual  & 8  &  20 &   31 \\
multiview  & 5  &  21  &  38\\
repeated & 10  &  22  &  40 \\
classification  &  5  &   27  &   44 \\
manipulation  &  6   &  28   &  43\\
simple-pick  &  4   &  42   &  47\\
\hline
all tasks & 30 & 22 & 37\\
\end{tabular}
\caption{ The task tags, and the averaged performance of the tasks containing each tag across all models.
\label{tab:task_tags}
}
\end{table}

To understand the factors in the tasks that influence the model performance, we designed a tag system to label the properties of the tasks. The tags for each task can be seen in \url{https://robochallenge.ai/benchmark_detail}. We correlate the task tags with the averaged performance of the models, and list them in Tab.~\ref{tab:task_tags}.

Here we give a description of the semantics of the tags: 
\begin{itemize}
\item \texttt{temporal}: Identical images may be received on different stages of the task;
\item \texttt{softbody}: Involving deformable objects;
\item \texttt{precise3d}: Required to grab or place the object at a precise location;
\item \texttt{bimanual}: Required to use both arms at once;
\item \texttt{multivew}: Required to use more than one cameras;
\item \texttt{repeated}: Repeating a skill at least three times;
\item \texttt{classification}: Different objects need to go to different locations;
\item \texttt{manipulation}: Involving hinges or racks;
\item \texttt{simple-pick}: Simple pick-and-place task.
\end{itemize}

From the table, we see that the \texttt{temporal} dependence and engagement of \texttt{softbody} is destroying the success rates. Because all of the models we tested are single-frame models, they hardly complete a full temporal task. Also, softbodies deform in an unpredictable way and usually require very fine localization of the grabbing point, making the models hard to deal with.

Tasks with a \texttt{precise3d} tag is noticeably harder (18\% with the tag, 25\% without the tag). The models all work at a low resolution of 224x224, so this is in line with what we would expect.

Contrary to our expectation, factors like \texttt{bimanual}, \texttt{multiview} or \texttt{repeated} do not create an additional decrease in their average performance. We argue that the ``global average'' task is not easy by design, as most tasks contain at least one type of difficulty. So, this reflects the fact that these factors are roughly the same level of hardness as the models.

The \texttt{classification} and \texttt{manipulation} tag has slightly higher success rates than the global average. Due to the capacity of the models, it should be expected that the ``semantic problems'' in the tasks are easy to solve.

In the task set, we purposely include some simple pick-and-place tasks, labeled \texttt{simple-pick}, as the easiest portion of the benchmark. As seen in the table, they indeed receive higher scores, about two times the global average success rates. For strong models like $\pi_{0.5}$, the success rate goes as high as 90\%, marking these tasks largely solved.

\subsection{Conclusion}

After analyzing the results, we see that the strength of the models differ considerably. The latest model $\pi_{0.5}$ from Physical Intelligence is remarkably stronger in almost all respects. Meanwhile, there are still factors that are inherently hard for VLAs, and we would expect them to be solved by future models.

{\small
\bibliographystyle{ieeenat_fullname}
\bibliography{11_references}
}

\appendix 
\clearpage
\section{Walkthrough of Submitting a Model for Evaluation}
\label{sec:appendix_section}

\begin{figure}[h]
\centering
\includegraphics[width=0.99\linewidth]{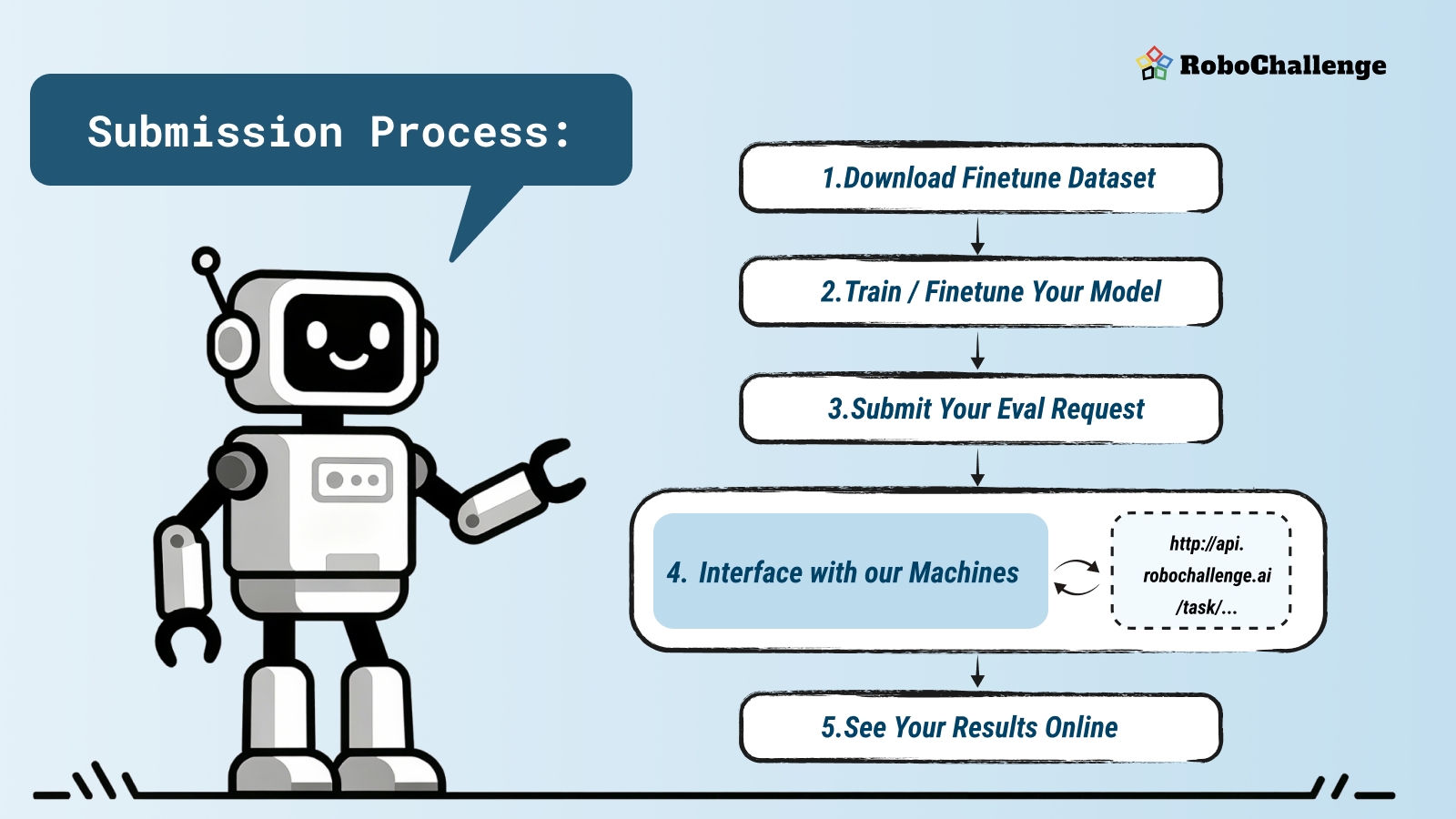}
\caption{
Intended workflow for participants.\label{fig:flow}
}
\end{figure}

Fig.~\ref{fig:flow} shows the intended workflow of a participant submitting their models to our benchmark.

Firstly, they need to download the demonstration dataset of the tasks. The data is hosted on Hugging Face. We release the data in a plain format, with video files and associating json format robot states stored separately. We also provide a utility script to convert the data to LeRobot format.

Next, they should decide on a setting and fine-tune their model. If the chosen setting is Generalist, multiple tasks should be trained at once, using the prompt to differentiate between tasks. If the chosen setting is Finetuned, there is no restriction on how the model is trained.

If some of the models are trained from the same basemodel, or if they use essentially the same algorithm except the difference in fine-tuning data, these models can share a common ``displayed name''. When we rank the algorithms for a benchmark, the results from the same user with the same displayed name will be grouped as one entry. Thus, the user can start from the same foundation model (e.g. $\pi_0$) and generate a task-specific finetuned model, and as long as they are submitted with a shared displayed name, the results will be ranked as one algorithm.

After training, the participants need to prepare for their submission. They need to figure out how to connect our API with the model inference code. To make this easier, we provide skeleton code to demonstrate how to interact with our API. Our skeleton code implements an observe-inference-stop cycle. Before model loading, the script regularly polls to see if the evaluation job is about to start. Minutes before the actual evaluation, the program gets noticed to prepare its weights, allocate memory on GPU, and warm-up the inference engine. During the evaluation, observations are retrieved, fed into the model, and actions are sent. The program will wait for the action queue to be cleared before the next request of the observations. This ensures that the images are captured in a steady state. In addition to the skeleton code, we also provide a mock test for the user to check that their code actually works.

After preparing the program, the user will be able to submit their evaluation request. In the submission, the user needs to provide its key, the desired task set, and the claimed model name. If multiple tasks are selected, the model will be assumed to be a multi-task generalist model.

When the evaluation request is submitted, it will be manually queued and scheduled. Our testing site is responsible for preparing all the props, setting up the case and recording the run. Because we need to make sure all the materials and the tester are ready, the wait time may be hours to days.

After the evaluation is completed, the resulting numbers and videos are available on the website. The user can view the machine logs in the RRD format using an open source viewer rerun.io.

By default, we open all the results of all participants to each other. One can view the recorded video of other's models to gain insight. If there is grading error on one's own model, the participant can contact us for re-calculation of the result.

\section{Photos of the Robot Platforms}
To give a more intuitive view of the robot platforms we are using, we list the example photo of each machine.

\noindent The UR5:
\begin{center}
\includegraphics[width=1.0\linewidth]{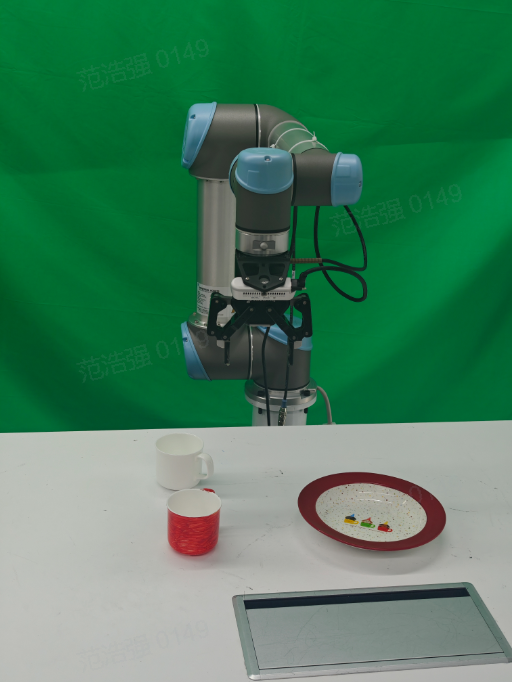}
\end{center}

The Cobot Magic ALOHA:
\begin{center}
\includegraphics[width=0.95\linewidth]{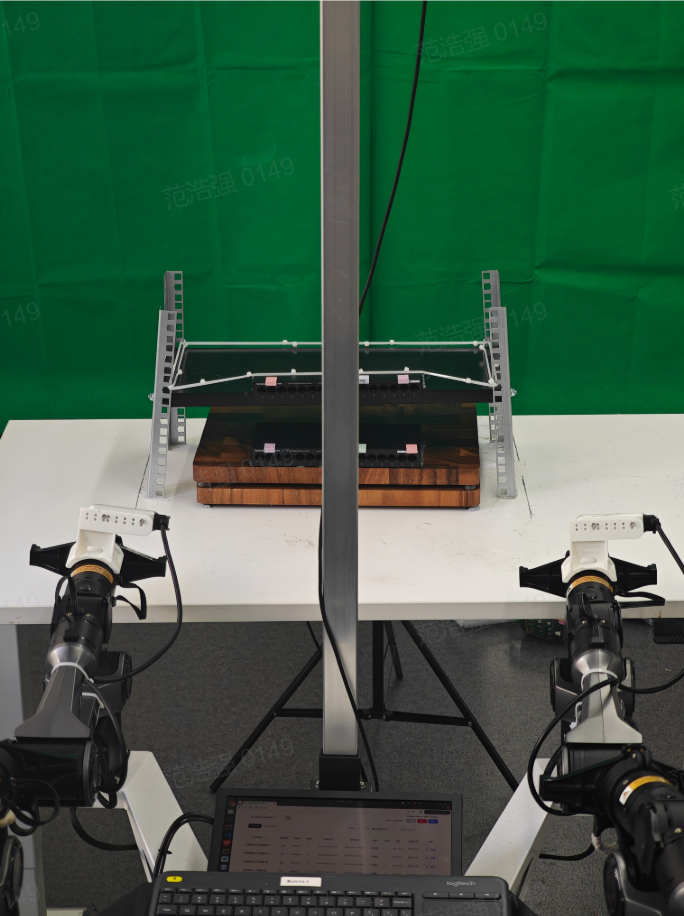}
\end{center}

The Franka Research 3:
\begin{center}
\includegraphics[width=0.95\linewidth]{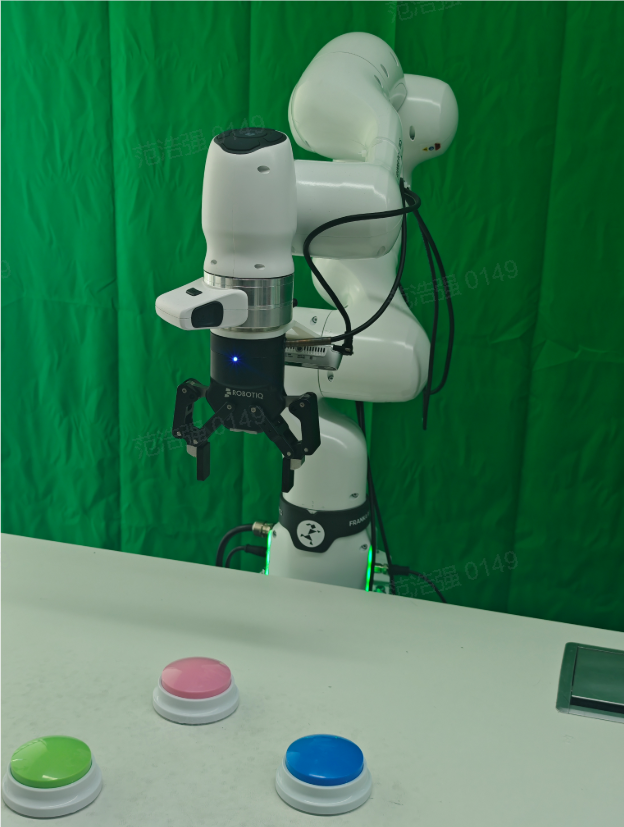}
\end{center}

The ARX5:
\begin{center}
\includegraphics[width=0.95\linewidth]{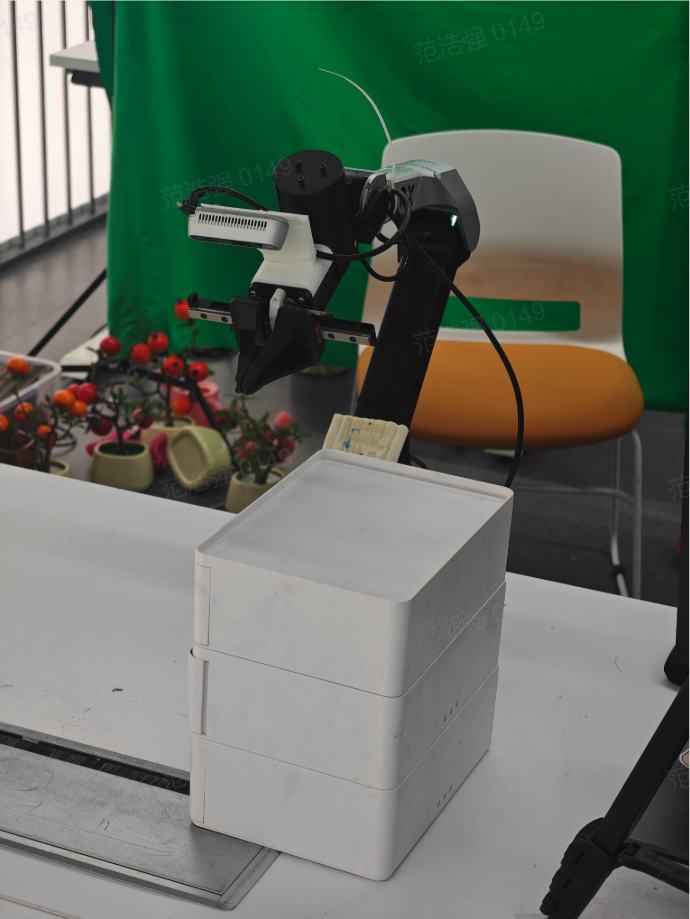}
\end{center}

\end{document}